\documentclass[12pt]{article}

\usepackage{amsmath}
\usepackage{tikz-cd}
\usepackage{amssymb}
\usepackage{graphicx}
\usepackage{hyperref}
\usepackage{geometry}
\usepackage{mathtools}
\usepackage{titlesec}
\usepackage{titling}
\usepackage[backend=bibtex,style=numeric,sorting=none,defernumbers=true]{biblatex}
\usepackage{etoolbox}   

\titleformat{\section}      {\normalfont\Large\bfseries\centering}{\thesection}{1em}{}
\titleformat{\subsection}   {\normalfont\large\bfseries\centering}{\thesubsection}{1em}{}
\titleformat{\subsubsection}{\normalfont\normalsize\bfseries\centering}{\thesubsubsection}{1em}{}

\titleclass{\subsubsubsection}{straight}[\subsubsection]
\newcounter{subsubsubsection}[subsubsection]
\renewcommand\thesubsubsubsection{\thesubsubsection.\arabic{subsubsubsection}}
\titleformat{\subsubsubsection}
  {\normalfont\normalsize\bfseries\centering}{\thesubsubsubsection}{1em}{}
\titlespacing*{\subsubsubsection}{0pt}{1.5ex plus 0.5ex minus .2ex}{1em}

\pretitle{%
  \begin{center}
    \vspace*{-0.42cm}
    \rule{\linewidth}{3pt}\\[\smallskipamount]
    \bfseries\LARGE
    \vspace{0.22cm}
}
\posttitle{%
    \\[\smallskipamount]
    \rule{\linewidth}{2pt}
  \end{center}
  \vspace{0.12cm}
}

\preauthor{\begin{center}\large}
\postauthor{\end{center}}

\predate{\begin{center}\large}
\postdate{\end{center}}

\makeatletter
\g@addto@macro\maketitle{\thispagestyle{empty}}
\makeatother

\title{Tensor Gauge Flow Models}

\author{%
  \begin{minipage}[t]{0.46\textwidth}\centering
    Alexander Strunk\thanks{Corresponding author: \href{mailto:astrunk.research@evercot.ai}{astrunk.research@evercot.ai}}\\
    Evercot AI\\
    \texttt{astrunk.research@evercot.ai}
  \end{minipage}%
  \hspace{1em}
  \begin{minipage}[t]{0.46\textwidth}\centering
    Roland Assam\\
    Evercot AI\\
    \texttt{rassam.research@evercot.ai}
  \end{minipage}%
\vspace{0.42cm}
}

\date{18 November 2025}

\begin{document}

\maketitle  

\vspace{-0.80cm} 
\begin{abstract}
\noindent
This paper introduces Tensor Gauge Flow Models, a new class of Generative Flow Models that generalize Gauge Flow Models \cite{GaugeFlowModels} and Higher Gauge Flow Models \cite{HigherGaugeFlowModels} by incorporating higher-order Tensor Gauge Fields into the Flow Equation. This extension allows the model to encode richer geometric and gauge-theoretic structure in the data, leading to more expressive flow dynamics.  Experiments on Gaussian mixture models show that Tensor Gauge Flow Models achieve improved generative performance compared to both standard and gauge flow baselines.
\end{abstract}

\newpage
\setcounter{page}{1}

\section{Introduction}
A Tensor Gauge Flow Model is a generative modeling approach in which the dynamics are governed by the neural ODE:
\begin{align*}
\hat{\nabla}_{dt} x(t)
=\; v_{\theta}\bigl(x(t),t\bigr)
\;-\; \alpha(x(t),t)\,\Pi_{M}\!\left(
\mathcal{A}_{\mu_{1}\dots \mu_{n}}\bigl(x(t),t\bigr)\!\left[\hat{T}\bigl(x(t),t\bigr)\right]\,
\prod_{i=1}^{n} d^{\mu_i}\!\bigl(x(t),t\bigr)
\right)
\end{align*}
with the following components:
\begin{itemize}
\item $v_{\theta}(x(t),  t) \in T_{x}M$ is a learnable vector field modeled by a Neural Network.
\item $\alpha (x(t), t)$ is a space and time dependent weight,  also modeled by a Neural Network.
\item  $\mathcal{A}_{\mu_{1}\dots \mu_{n}}(x(t),t)$ is the Tensor Gauge Field modeled by a Neural Network.
\item $d^{\mu}(x(t), t) \in T_{x}M$ is the direction vector field.
\item  $\hat{T}(x(t),t)$ is a Tensor Field on $C$,  which can be learned by a Neural Network.
\item $\Pi_{M}: C \rightarrow TM$ is a smooth projection from the vector bundle $C$ to the tangent bundle $TM$ of the base manifold $M$.  
\end{itemize}

\noindent The action of the Tensor Gauge Field $\mathcal{A}_{\mu_{1}\dots \mu_{n}}(x(t),t)$ can be either:
\begin{itemize}
  \item \textbf{Action of a Lie algebra \(\mathfrak{g}\) on the Tensor Field \(\hat{T}(x(t),t)\):}
  \begin{align*}
    \mathcal{A}_{\mu_{1}\dots \mu_{n}}\bigl(x(t),t\bigr)\!\left[\hat{T}\bigl(x(t),t\bigr)\right]
    :=\;
    \mathcal{A}^{a}_{\mu_{1}\dots \mu_{n}}\bigl(x(t),t\bigr)\,
    L_{a}\!\left[\hat{T}\bigl(x(t),t\bigr)\right],
  \end{align*}
  where $\{L_a\}$ are the generators of $\mathfrak{g}$ in the representation carried by $\hat{T}(x(t),t)$.  
  The tensor field may be an ordinary vector, $\hat{T}^{\alpha}(x(t),t)=\hat{v}^{\alpha}(x(t),t)$, or a contraction
  \[
    \hat{T}^{\alpha}(x(t),t)
    = \sum_{k} T^{\alpha}{}_{\nu_{1}\dots \nu_{k}}(x(t),t)\,
      \hat{v}^{\nu_{1}}(x(t),t)\cdots \hat{v}^{\nu_{k}}(x(t),t).
  \]

  \item \textbf{Action of an \(L_{\infty}\)-algebra on the (graded) Tensor Field \(\hat{T}(x(t),t)\):}
  \begin{align*}
    \mathcal{A}_{\mu_{1}\dots \mu_{n}}\bigl(x(t),t\bigr)\!\left[\hat{T}\bigl(x(t),t\bigr)\right]
    :=\;
    \sum_{m}
    \mathcal{A}^{a}_{\mu_{1}\dots \mu_{n}}\bigl(x(t),t\bigr)\,
    b_{m}\!\bigl(e_{a},\,\hat{T}\bigl(x(t),t\bigr),\dots,\hat{T}\bigl(x(t),t\bigr)\bigr),
  \end{align*}
  where $b_m$ are the higher multilinear brackets of the $L_{\infty}$-algebra and $\{e_a\}$ is a chosen basis.  
  Again, $\hat{T}(x(t),t)$ may be a graded vector $\hat{v}$ or a graded contracted tensor as above.
\end{itemize}

 \newpage
\section{Mathematical Background}
In this section the mathematical background foundational to Tensor Gauge Flow Models is introduced. The central framework is that of Tensor Fields on Fiber Bundles. Standard references for these Differential-Geometric notions include \cite{DifferentialGeometry, DifferentialGeometryManifoldConnections, DifferentialGeometryManifold, SmoothManifolds, Manifolds, MathematicalGaugeTheory, GeometryTopologyPhysics, FibreBundle}. For an intuitive and concise exposition adapted to the present setting, the construction builds on the mathematical concepts described in \cite{GaugeFlowModels, HigherGaugeFlowModels}.

\subsection{Tensor Gauge Fields}
A Tensor Gauge Field is a Tensor Field   $\mathcal{A}^{\mu_{1}\dots\mu_{m}}{}_{\nu_{1}\dots\nu_{n}}{}^{a}$  whose coefficients take values in a (graded) vector space carrying either an $L_{\infty}$-algebra or a Lie-algebra structure. Equivalently, if $E\to M$ is a (graded) vector bundle with typical fiber $V$, a Tensor Gauge Field on $M$ is a section of $T^{(m,n)}M\otimes E$.
Here,  the fiber index ${}^a$ refers to a choice of local frame of $E$.
 A Tensor Gauge Flow Model can be visualized locally as follows:
\begin{center}
\begin{figure}[h!]
  \centering
  \includegraphics[width=0.55\textwidth]{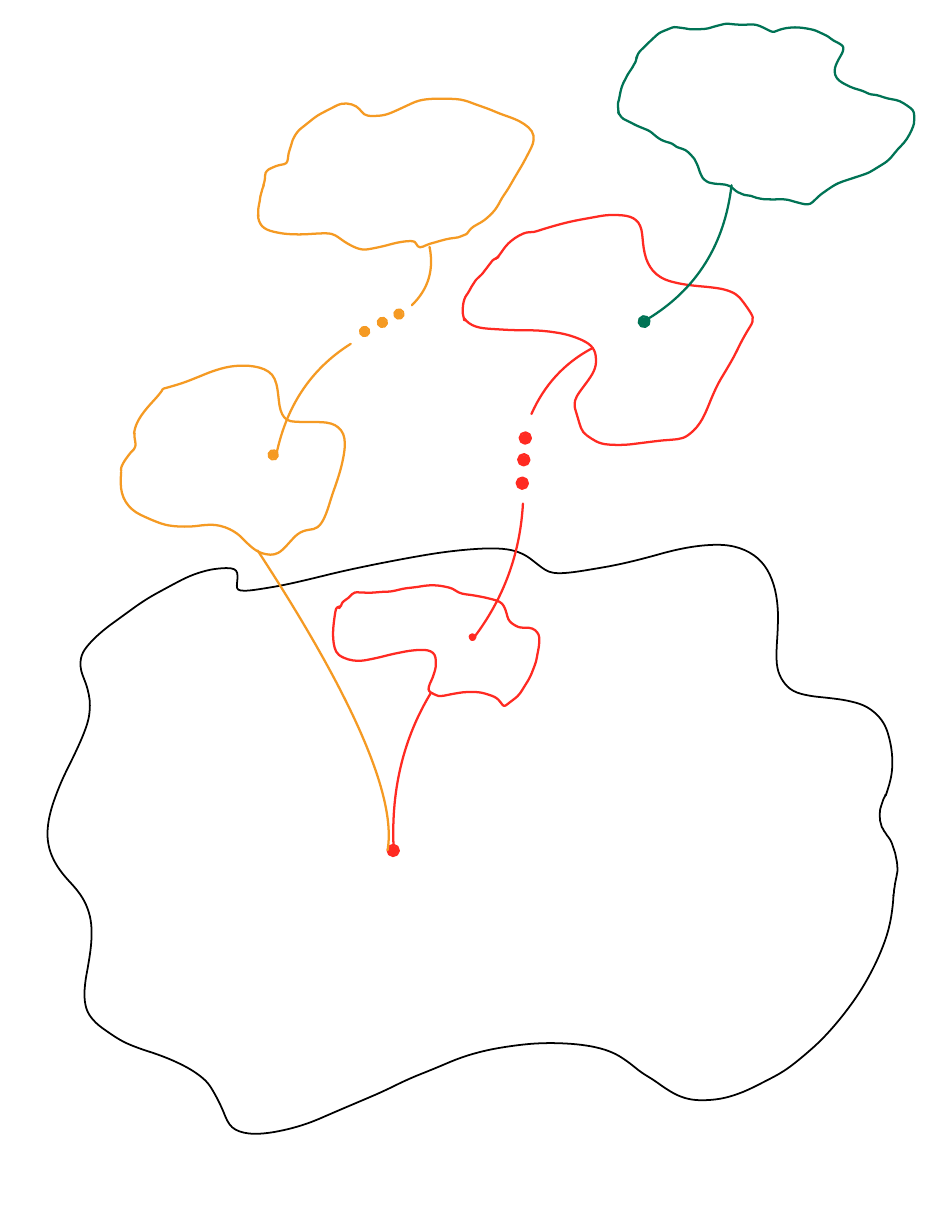}
  \caption{A Tensor Gauge Flow Model  can be visualized locally via its Tensor Gauge Field $\mathcal{A}^{\mu_{1}\dots\mu_{m}}{}_{\nu_{1}\dots\nu_{n}}{}^{a}$. The black region represents the base manifold $M$ on which the trajectory $x(t)$ evolves.  Above each point of $M$, the red regions depict the fiber spaces associated with the indices $\mu_{1},\dots,\mu_{m}$ while the green region represents the internal vector space carrying the index $a$.
The yellow regions illustrate the Tensor Field  $\hat{T}(x(t),t)$ evaluated along the flow,  which is acted on by $\mathcal{A}$ to produce the Tensor Gauge correction in the dynamics.}
  \label{fig:TGFM}
\end{figure}
\end{center}
\newpage
\subsubsection{Tensor Fields on Fiber Bundles}
In mathematical terms,  an $(m,n)$-Tensor Field $\hat{T}^{(m,n)}$ on a manifold $M$ is a smooth section:
\[
\hat{T}^{(m,n)} \in \bigl\{\, S : M \rightarrow T^{(m,n)}M \ \big|\ \pi \circ S = \mathrm{Id}_{M} \,\bigr\},
\]
where the $(m,n)$-Tensor bundle is
\[
T^{(m,n)}M := \bigsqcup_{x \in M} T^{(m,n)}_{x}M,
\]
with projection $\pi: T^{(m,n)}M \rightarrow M$, and fibers
\[
T^{(m,n)}_{x}M := (T_{x}M)^{\otimes m} \otimes (T^{*}_{x}M)^{\otimes n}.
\]
In local coordinates $x^{\mu}$, a Tensor Field has the form
\[
\hat{T}^{(m,n)}
= \hat{T}^{\mu_{1}\dots\mu_{m}}{}_{\nu_{1}\dots\nu_{n}}\,
\frac{\partial}{\partial x^{\mu_{1}}}
\otimes \cdots \otimes
\frac{\partial}{\partial x^{\mu_{m}}}
\otimes dx^{\nu_{1}} \otimes \cdots \otimes dx^{\nu_{n}}.
\]
\noindent A (smooth) Fiber Bundle is a surjective map $\pi:E\rightarrow M$ with typical fiber $F$ such that for every $x\in M$ there is an open neighborhood $U$ and a diffeomorphism $\phi_U:\pi^{-1}(U)\!\to U\times F$ satisfying $\mathrm{pr}_1\circ \phi_U=\pi$ (local triviality).\\
 There are now two possibilities to express a Tensor Gauge Field on a Fiber Bundle $E \rightarrow M$:
\begin{enumerate}
\item \emph{Tensor Field on the base manifold valued in the fiber:}
\[
\hat{T}^{(m,n)}  \;\in\; \bigl\{\, S : M \rightarrow T^{(m,n)}M \otimes E \ \big|\ \pi_{T^{(m,n)}M \otimes E} \circ S = \mathrm{Id}_{M} \,\bigr\},
\]
valued in the typical fiber $F$ (assuming $E$ is a vector bundle, e.g., an associated bundle).
\item \emph{Tensor Field on the total space:}
\[
\hat{T}^{(m,n)} \;\in\; \bigl\{\, S : E \rightarrow T^{(m,n)}E \ \big|\ \pi_{T^{(m,n)}E} \circ S = \mathrm{Id}_{E} \,\bigr\}.
\]
\end{enumerate}

\newpage

\section{Tensor Gauge Flow Models}
A Tensor Gauge Flow Model is defined on a (graded) Fiber Product Bundle $\hat{A}$:
\[
\hat{A} = B \times_{M} C
\]
where $B \to M$ and $C \to M$ are smooth Fiber Bundles over the same base: 
\begin{center}
\begin{tikzcd}[column sep=3.5em,row sep=3.5em]
& {\scalebox{1.5}{$\substack{\hat{A}\\=\\B\times_M C}$}}
    \arrow[dl, swap, "p_B"]
    \arrow[dr, "p_C"]
& \\[-0.2em]
B
    \arrow[dr, swap, "\pi_B"]
& 
& C
    \arrow[dl, "\pi_C"] \\[0.2em]
& M &
\end{tikzcd}
\end{center}
The dynamics of the Tensor Gauge Flow Model is governed by the following ODE:
\begin{align*}
\hat{\nabla}_{dt} x(t)
=\; v_{\theta}\bigl(x(t),t\bigr)
\;-\; \alpha(x(t),t)\,\Pi_{M}\!\left(
\mathcal{A}_{\mu_{1}\dots \mu_{n}}\bigl(x(t),t\bigr)\!\left[\hat{T}\bigl(x(t),t\bigr)\right]\,
\prod_{i=1}^{n} d^{\mu_i}\!\bigl(x(t),t\bigr)
\right)
\end{align*}
with the following ingredients:
\begin{itemize}
\item $v_{\theta}(x(t),  t) \in TM$ represents a learnable vector field modeled by a Neural Network.
\item $\alpha (x(t), t)$ is a space-and time-dependent scalar weight,  also modeled by a Neural Network.
\item  $\mathcal{A}_{\mu_{1}\dots \mu_{n}}(x(t),t)$ is the Tensor Gauge Field modeled by a Neural Network.
\item $d^{\mu}(x(t), t) \in TM$ is the direction vector field.
\item  $\hat{T}(x(t),t)$ is a Tensor Field on $C$,  which can be parameterized by a Neural Network.
\item $\Pi_{M}: C \rightarrow TM$ is a smooth bundle map projection from the vector bundle $C$ to the tangent bundle $TM$ of the base manifold $M$.  
\end{itemize}

\noindent The Tensor Gauge Field $\mathcal{A}_{\mu_{1}\dots \mu_{n}}(x(t),t)$ is encoded by the Fiber Bundle $B$.  The fibers of $B$ are different for the following two cases:
\begin{itemize}
  \item \textbf{$L_{\infty}$-algebra-valued case:} The fibers of $B$ carry a graded vector space $\hat{B}$ underlying an L$_{\infty}$-algebra,  and $\mathcal{A}$ is a smooth section of the tensor bundle $T^{(m,n)}M$ with values in $\hat{B}$:
\[
\mathcal{A}_{\mu_{1}\dots \mu_{n}}(x(t),t) \in \{\mathcal{A}: M \to T^{(m,n)}M\otimes \hat{B},\ \pi_{T^{(m,n)}M\otimes \hat{B}}\circ \mathcal{A}=\mathrm{Id}_M \}
\]
\item \textbf{Lie-algebra-valued case:} The fibers of $B$ carry a Lie algebra $\mathfrak{g}$:
\[
\mathcal{A}_{\mu_{1}\dots \mu_{n}}(x(t),t) \in \{\mathcal{A}: M \to T^{(m,n)}M\otimes \mathfrak{g},\ \pi_{T^{(m,n)}M\otimes \mathfrak{g}}\circ \mathcal{A}=\mathrm{Id}_M \}
\]
\end{itemize}
\noindent The corresponding Tensor Field $\hat{T}(x(t),t)$ is a Tensor Field on $C$ with typical fiber $P$, such that:
\begin{itemize}
\item  \textbf{$L_{\infty}$-algebra-valued case:} $\hat{T}(x(t),t)$ takes values in a graded fiber $\hat{P}$,  which is equipped fibrewise with the same L$_{\infty}$-algebra structure as $\hat{B}$:
\[
\hat{T}(x(t),t) \in \{\hat{T}: M \to T^{(m,n)}M\otimes \hat{P},\ \pi_{T^{(m,n)}M\otimes \hat{P}}\circ \hat{T}=\mathrm{Id}_M \}
\]
\item \textbf{Lie-algebra representation case:} $\hat{T}(x(t),t)$ takes values in a representation $P$ of $\mathfrak{g}$:
\[
\hat{T}(x(t),t \in \{\hat{T}: M \to T^{(m,n)}M\otimes P,\ \pi_{T^{(m,n)}M\otimes P}\circ \hat{T}=\mathrm{Id}_M \}
\]
\end{itemize}
The action of the Tensor Gauge Field $\mathcal{A}$ on the Tensor Field $\hat{T}$ differs in the two cases:
\begin{itemize}
\item \textbf{Action of a Lie algebra \(\mathfrak{g}\) on the Tensor Field \(\hat{T}(x(t),t)\):}
\begin{align*}
\mathcal{A}_{\mu_{1}\dots \mu_{n}}\bigl(x(t),t\bigr)\!\left[\hat{T}\bigl(x(t),t\bigr)\right]
:=\;
\mathcal{A}^{a}_{\mu_{1}\dots \mu_{n}}\bigl(x(t),t\bigr)\,L_{a}\!\left[\hat{T}\bigl(x(t),t\bigr)\right],
\end{align*}
with $L_a$ the basis elements of $\mathfrak{g}$ acting in the representation carried by $\hat{T}(x(t),t)$.  
The Tensor Field may be an ordinary vector, \(\hat{T}^{\alpha}(x(t),t)=\hat{v}^{\alpha}(x(t),t)\), or a contraction
\(\hat{T}^{\alpha}(x(t),t)=\sum_{k} T^{\alpha}_{\nu_{1}\dots \nu_{k}}\,\hat{v}^{\nu_{1}}\!\cdots \hat{v}^{\nu_{k}}\).

\item \textbf{Action of an \(L_{\infty}\)-algebra on the (graded) Tensor Field \(\hat{T}(x(t),t)\):}
\begin{align*}
\mathcal{A}_{\mu_{1}\dots \mu_{n}}\bigl(x(t),t\bigr)\!\left[\hat{T}\bigl(x(t),t\bigr)\right]
:=\;
\sum_{m}\mathcal{A}^{a}_{\mu_{1}\dots \mu_{n}}\bigl(x(t),t\bigr)\,
b_{m}\!\bigl(e_{a},\,\hat{T}\bigl(x(t),t\bigr),\dots,\hat{T}\bigl(x(t),t\bigr)\bigr),
\end{align*}
where \(b_m\) are the higher multilinear brackets and \(e_a\) a basis.  
Again, \(\hat{T}(x(t),t)\) may be a graded Vector \(\hat{v}\) or a graded contracted Tensor as above.
\end{itemize}

\medskip
\noindent
The Tensor Gauge Flow Model (TGFM) is defined for any smooth base manifold $M$.  As in ordinary Higher Gauge Flow Models~\cite{HigherGaugeFlowModels}, training a TGFM requires $M$ to carry a Riemannian metric $g$, so that $M$ becomes a Riemannian manifold.  The training procedure is based on the Riemannian Flow Matching (RFM) framework~\cite{RFM}, which generalizes Flow Matching (FM)~\cite{FM}.  The TGFM loss is given by
\[
\mathcal{L}_{\mathrm{TGFM}}
=  \mathbb{E}_{\substack{t \sim \mathcal{U}[0,1] \\ x \sim p_t}}
  \bigg\lVert
    \Big[
      v_{\theta}\bigl(x,t\bigr)
      \;-\; \alpha(x(t),t)\,\Pi_{M}\!\left(
        \mathcal{A}_{\mu_{1}\dots \mu_{n}}\bigl(x,t\bigr)\!\left[\hat{T}\bigl(x,t\bigr)\right]\,
        \prod_{i=1}^{n} d^{\mu_i}\!\bigl(x,t\bigr)
      \right)
    \Big]
    - u_{t}(x)
  \bigg\rVert_{g_x}^{2},
\]
where $\|\cdot\|_{g_x}$ is the norm induced by the Riemannian metric $g$ at $x \in M$.  Both terms inside the norm are tangent vectors at $x$:
\begin{gather*}
\Big[
  v_{\theta}\bigl(x,t\bigr)
  \;-\; \alpha(x(t),t)\,\Pi_{M}\!\left(
    \mathcal{A}_{\mu_{1}\dots \mu_{n}}\bigl(x,t\bigr)\!\left[\hat{T}\bigl(x,t\bigr)\right]\,
    \prod_{i=1}^{n} d^{\mu_i}\!\bigl(x,t\bigr)
  \right)
\Big] \in T_x M,\\[0.3em]
u_{t}(x) \in T_x M.
\end{gather*}

\noindent
The TGFM loss incorporates the following components:
\begin{itemize}
  \item \textbf{Probability-density path:}
    a smooth family of probability densities
    $p_t : M \to \mathbb{R}_{+}$, $t \in [0,1]$, as in~\cite{RFM}, such that for each $t$
    \[
      \int_{M} p_t(x)\, d\mathrm{vol}_x = 1,
    \]
    where $d\mathrm{vol}_x$ is the Riemannian volume form induced by $g$.

  \item \textbf{Target vector field:}
    a marginal target field $u_t(x)$ obtained from a conditional field $u_t(x \mid x_1)$ by
    \[
      u_t(x) =
      \int_{M}
        u_t\!\bigl(x \mid x_1\bigr)\,
        \frac{p_t\!\bigl(x \mid x_1\bigr)\, q(x_1)}{p_t(x)}
      \, d\mathrm{vol}_{x_1},
    \]
    where $q(x) \coloneqq p_{t=1}(x)$ and $x_1$ denotes the sample at time $t=1$.
\end{itemize}

\noindent
In general, evaluating $\mathcal{L}_{\mathrm{TGFM}}$ exactly is computationally intractable, as it requires integrating over the marginal target field $u_t(x)$.  By construction, this loss coincides with the Riemannian Flow Matching (RFM) objective.  To obtain a tractable training criterion,~\cite{RFM} introduces the Riemannian Conditional Flow Matching (RCFM) loss, an unbiased single-sample Monte Carlo estimator of the RFM objective.  The key idea is to replace the intractable marginal field $u_t(x)$ by its conditional counterpart $u_t(x \mid x_1)$, enabling simulation-free learning on simple geometries and scalable neural network training on general Riemannian manifolds.  The same construction applies directly to Tensor and Higher Gauge Flow Models.

\noindent
For detailed derivations and implementation specifics, see~\cite{RFM,FM}.

\newpage

\section{Experiments}

This section empirically compares several variants of Tensor Gauge Flow Models (TGFM) against a plain Flow Matching baseline and a standard Gauge Flow Model on synthetic Gaussian mixture data in ambient dimension \(N\).  
Throughout, the base manifold is
\[
M \;=\; \mathbb{R}^{N},
\]
equipped with the standard Euclidean metric \(g\), so that \(T_x M \cong \mathbb{R}^N\) and
\(\|\cdot\|_{g_x}\) is the usual \(\ell_2\)-norm.  
The bundles \(B \to M\) and \(C \to M\) appearing in the definition of TGFM are taken to be trivial:
\[
B \;\cong\; M \times \hat{B},
\qquad
C \;\cong\; M \times P,
\]
and the experiments are restricted to the Lie-algebra-valued case with
\[
\mathfrak{g} \;=\; \mathfrak{so}(N),
\qquad
P \;=\; \mathbb{R}^N,
\]
equipped with the defining representation of \(\mathfrak{so}(N)\).
All models are trained with the Euclidean specialization of the TGFM/Riemannian Flow Matching
objective introduced in Section~3, and differ only in the neural parameterization of the
vector field \(v_{\theta}\) and the Tensor Gauge Field \(\mathcal{A}\).
The total number of trainable parameters is approximately matched across model families.

\subsection{TGFM Dynamics and Gauge Correction}

On the Euclidean manifold \(M = \mathbb{R}^N\) the covariant derivative
\(\hat{\nabla}_{dt} x(t)\) reduces to the ordinary time derivative \(\tfrac{d}{dt}x(t)\),
and the TGFM dynamics from Section~3 take the form
\[
\frac{d}{dt}x(t)
=\; v_{\theta}\bigl(x(t),t\bigr)
\;-\; \alpha\bigl(x(t),t\bigr)\,
\Pi_{M}\!\left(
\mathcal{A}_{\mu_{1}\dots \mu_{n}}\bigl(x(t),t\bigr)\!\left[
\hat{T}\bigl(x(t),t\bigr)
\right]
\prod_{i=1}^{n} d^{\mu_i}\!\bigl(x(t),t\bigr)
\right),
\]
where \(\Pi_M : C \to TM\) is the projection from the auxiliary bundle \(C\) to the tangent
bundle \(TM\).  In all experiments, the auxiliary bundle is chosen as \(C = TM\), so that
\(\Pi_M\) is the identity, and the highest tensor rank is restricted to \(n = 3\), which yields
\[
\frac{d}{dt}x(t)
=\; v_{\theta}\bigl(x(t),t\bigr)
\;-\; \alpha\bigl(x(t),t\bigr)\,
\sum_{k=1}^{3}
\mathcal{A}_{\mu_{1}\dots \mu_{k}}\bigl(x(t),t\bigr)\!\left[
\hat{T}\bigl(x(t),t\bigr)
\right]
\prod_{i=1}^{k} d^{\mu_i}\!\bigl(x(t),t\bigr).
\]
The corresponding Tensor Field \(\hat{T}(x(t),t)\) is instantiated as a finite polynomial in
the base vector field \(v_{\theta}\):
\[
\hat{T}(x(t),t) \;=\; \sum_{k=1}^{2} T_{\mu_{1} \dots \mu_{k}}^{a}(x(t),t)\,
v^{\mu_{1}}_{\theta}(x(t),t)\cdots v^{\mu_{k}}_{\theta}(x(t),t),
\]
where the superscript \(a\) indexes a fixed basis of the representation space of
the Lie algebra \(\mathfrak{so}(N)\).

\subsubsection*{Model variants within the TGFM framework}

The following five model families are compared on the Euclidean manifold \((M,g)\):

\begin{itemize}
  \item \textbf{Plain Flow Model (PlainVF).}  
        A baseline Flow Model with no gauge correction.  
        The effective flow vector field is modeled by a single MLP,
        \[
        v_{\theta}^{\mathrm{eff}}(x,t)
        =
        v_{\theta}(x,t) : M\times[0,1]\to TM,
        \]
        implemented as a fully-connected network with SiLU nonlinearities.

  \item \textbf{Gauge Flow Model (PlainVF + Plain Gauge Field).}  
        A Lie-algebra-valued Gauge Flow Model with
        \[
        v_{\theta}^{\mathrm{aux}},\,v_{\theta}^{\mathrm{main}} : M\times[0,1] \to TM
        \]
        both realized as plain MLPs, and \(A\) parameterized by an MLP for
        \(a_{\theta}(x,t)\) followed by the canonical skew-symmetric basis
        of \(\mathfrak{so}(N)\).

  \item \textbf{TensorVF + Plain Gauge Field.}  
        As in the Gauge Flow Model, but \(v_{\theta}^{\mathrm{aux}}\) and \(v_{\theta}^{\mathrm{main}}\)
        are replaced by Tensor Vector Fields constructed via \(\hat{T}(x(t),t)\) of
        maximal rank \(2\), providing a tensorized realization of the
        TGFM vector and direction fields, while the Gauge Field remains a
        plain MLP-based \(\mathfrak{so}(N)\)-valued section.

  \item \textbf{PlainVF + Tensor Gauge Field.}  
        The vector fields \(v_{\theta}^{\mathrm{aux}}\) and \(v_{\theta}^{\mathrm{main}}\)
        are plain MLPs, but the Gauge Field \(\mathcal{A}\) is promoted to a
        Tensor Gauge Field of maximal rank \(3\).

  \item \textbf{TensorVF + Tensor Gauge Field.}  
        Both vector fields \(v_{\theta}^{\mathrm{aux}}\) and \(v_{\theta}^{\mathrm{main}}\) are 
        Tensor Vector Fields constructed from \(\hat{T}(x(t),t)\), and the Gauge Field
        \(\mathcal{A}\) is a Tensor Gauge Field of maximal rank \(3\).
\end{itemize}

\subsection{Dataset}

All experiments are conducted on synthetic Gaussian mixture data in \(\mathbb{R}^N\).  
For a given ambient dimension \(N\), a mixture of \(K\) isotropic Gaussian components is constructed as
\[
p(x) \;=\; \frac{1}{K} \sum_{k=0}^{K-1} \mathcal{N}\bigl(x; \mu_k, 0.5\,I_N\bigr),
\]
with equal mixture weights \(\pi_k = 1/K\) and covariance \(\Sigma_k = 0.5\,I_N\) for all \(k\).  
In the experiments, the configuration
\[
K = 10{,}000, \qquad \alpha =  250
\]
is used.

\paragraph{Component means.}
For each component index \(k \in \{0,\dots,K-1\}\), a mean vector \(\mu_k\in\mathbb{R}^N\) is constructed as follows:
\begin{enumerate}
  \item \emph{Primary axis:}  
        Set \(a_1 = k \bmod N\) and
        \[
        (\mu_k)_{a_1} = \alpha \cdot 
        \begin{cases}
        1, & \text{if } k \text{ is even},\\
        -1, & \text{otherwise}.
        \end{cases}
        \]
  \item \emph{Secondary axis:}  
        Let \(a_2 = (k + \lfloor K/2\rfloor) \bmod N\).  
        If \(a_2 \neq a_1\), define
        \[
        (\mu_k)_{a_2} 
        = 0.5\,\alpha \cdot 
        \begin{cases}
        1, & \text{if } (k+1) \text{ is even},\\
        -1, & \text{otherwise}.
        \end{cases}
        \]
  \item \emph{Additional offset for overcomplete mixtures:}  
        Whenever \(K > N\) and \(k \ge N\), a small dimension-dependent offset is added on a third coordinate.  
        Let \(b = (a_1 + \lfloor k/N\rfloor) \bmod N\) and
        \[
          (\mu_k)_b \;{+}= 
          \bigl(1 \text{ if } k \bmod 3 = 0, \text{ else } -1\bigr)\,
          0.1\,\alpha \,\lfloor k/N\rfloor.
        \]
\end{enumerate}
This construction yields a rich, anisotropic yet isotropically covariant mixture with well-separated modes whose layout systematically depends on both the component index \(k\) and the ambient dimension \(N\).

\paragraph{Sampling.}
To draw a data point, a component index \(k \sim \mathrm{Cat}(\pi)\) is sampled, followed by
\[
x \sim \mathcal{N}(\mu_k, 0.5\,I_N).
\]
For each ambient dimension \(N\), the dataset sizes are
\[
n_{\text{train}} = 45{,}000, \qquad 
n_{\text{test}} =  15{,}000.
\]

\subsection{Training Objective and Optimization}

\subsubsection*{Euclidean specialization of the TGFM loss}

As discussed in Section~3, the TGFM loss on \((M,g)=(\mathbb{R}^N,\text{Euclidean})\)
reduces to the standard Euclidean Flow Matching (FM) loss when the probability-density path
\(p_t\) and the conditional vector fields are chosen appropriately.
The usual Euclidean Conditional Flow Matching construction~\cite{FM} is adopted:
given a data point \(x_1 \sim q\) and an independent reference sample
\(x_0 \sim \mathcal{N}(0,I_N)\), a time \(t \sim \mathrm{Unif}(0,1)\) is sampled and the straight
line (geodesic) interpolation
\[
\hat{x}_t = t\,x_1 + (1-t)\,x_0, \qquad
u_t = x_1 - x_0
\]
is used.
For an effective TGFM velocity field \(v_{\theta}^{\mathrm{eff}}\),
the Euclidean Flow Matching loss is
\[
\mathcal{L}_{\text{FM}}(\theta)
  = \mathbb{E}\bigl[
      \|v_{\theta}^{\mathrm{eff}}(\hat{x}_t,t) - u_t\|_2^2
    \bigr],
\]
which is the Euclidean specialization of the Riemannian Flow Matching objective with
\(g_x\) equal to the identity metric.

\noindent For the gauge-based models this loss is augmented with two regularizers acting on the
Lie-algebra-valued Tensor Gauge Field:
\[
\mathcal{L}(\theta)
 =
 \mathcal{L}_{\text{FM}}(\theta)
 + \lambda_A\,\mathcal{L}_A(\theta)
 + \lambda_{\text{cons}}\,\mathcal{L}_{\text{cons}}(\theta),
\]
where:
\begin{itemize}
  \item \(\mathcal{L}_A(\theta)\) penalizes large Gauge Fields, implemented as the mean
        squared Frobenius norm of the \(\mathfrak{so}(N)\)-matrix section:
        \[
        \mathcal{L}_A(\theta)
          = \mathbb{E}\bigl[\|A_{\theta}(\hat{x}_t,t)\|_F^2\bigr],
        \]
  \item \(\mathcal{L}_{\text{cons}}(\theta)\) is a \emph{gauge consistency} regularizer
        that penalizes large spatial and temporal gradients of the Lie-algebra
        coefficients \(a_{\theta}(x,t)\):
        \[
        \mathcal{L}_{\text{cons}}(\theta)
        =
        \mathbb{E}\bigl[\,
           \|\nabla_x a_{\theta}(\hat{x}_t,t)\|_2
          +\|\partial_t a_{\theta}(\hat{x}_t,t)\|_2
        \,\bigr],
        \]
        implemented via automatic differentiation of \(\sum_i (a_{\theta})_i\) with
        respect to \(x\) and \(t\).
\end{itemize}
The regularization coefficients are fixed to
\[
\lambda_A = 10^{-5}, \qquad
\lambda_{\text{cons}} = 10^{-5},
\]

\subsubsection*{Optimization and training schedule}

For each dimension \(N\), all TGFM parameterizations are trained from scratch with the
same optimization hyperparameters:
\begin{itemize}
  \item optimizer: AdamW with learning rate \(10^{-4}\) and weight decay \(10^{-6}\),
  \item batch size: \(256\),
  \item number of epochs: \(180\),
  \item gradient clipping at global norm \(1.0\).
\end{itemize}

\subsection{Results}

Across all considered dimensions \(N\), the Tensor Gauge Flow Models consistently outperform both the plain Flow Models and the standard Gauge Flow Models.

\newpage

\subsubsection{Training loss}

\begin{figure}[h!]
  \centering
  \includegraphics[width=0.6\textwidth]{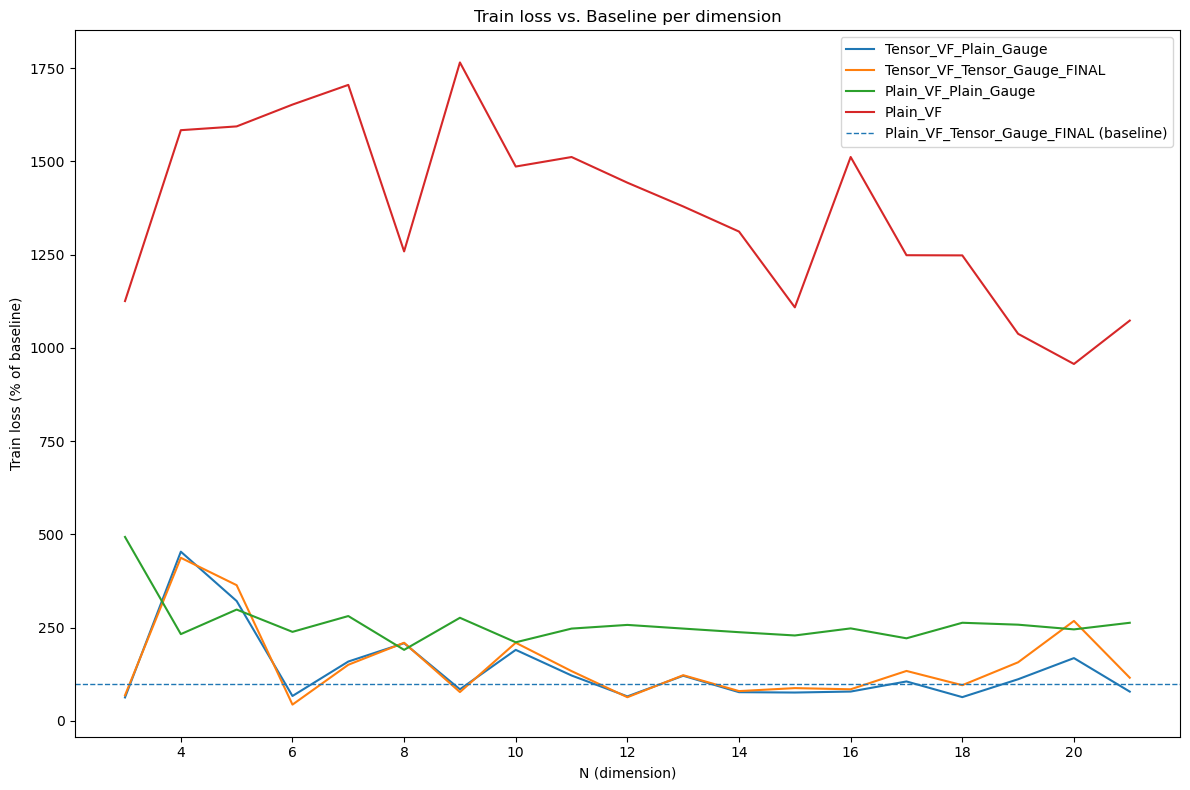}
  \caption{Training loss comparison (lower is better).  
  The training loss for each model is normalized by the loss of the PlainVF + Tensor Gauge Field model.  
  Values are shown for several ambient dimensions \(N\).}
  \label{fig:train_loss}
\end{figure}

Figure~\ref{fig:train_loss} compares the training losses of all model variants.  
The Tensor Gauge Flow Models achieve the lowest training loss across all dimensions \(N\).

\subsubsection{Test loss}

\begin{figure}[h!]
  \centering
  \includegraphics[width=0.6\textwidth]{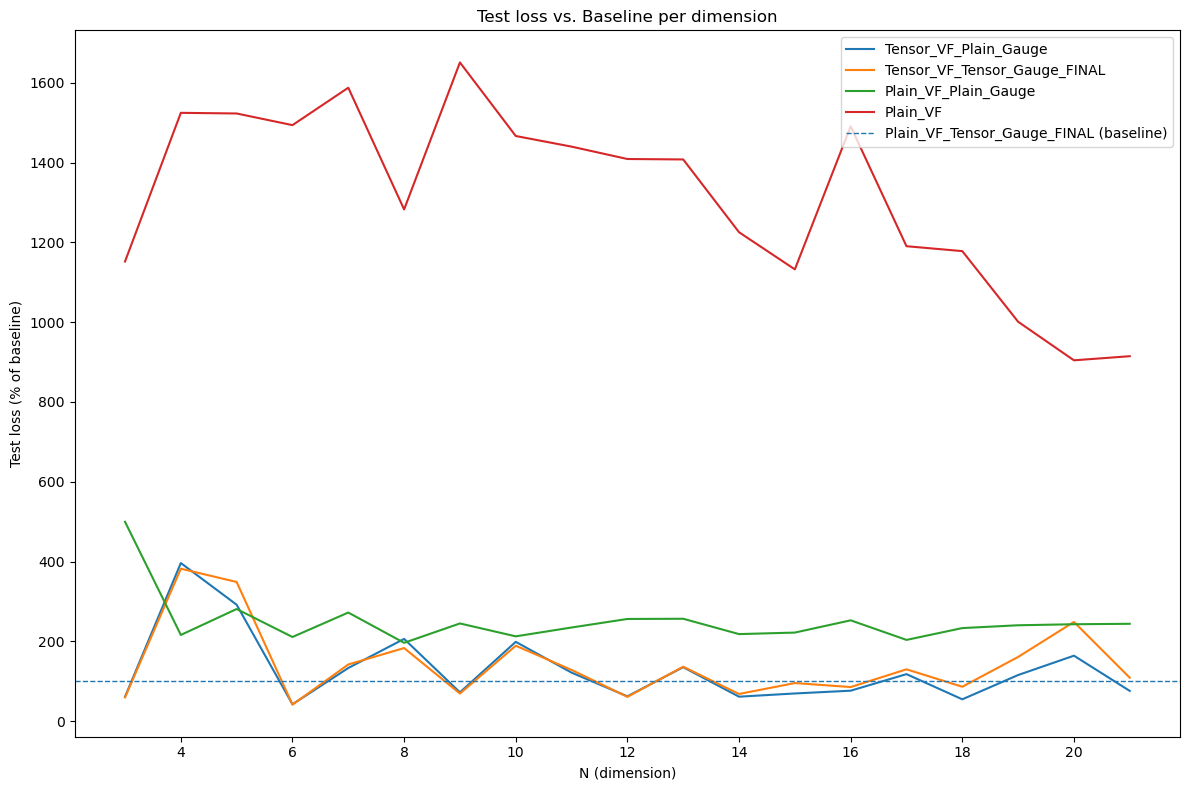}
  \caption{Test loss comparison (lower is better).  
  The test loss for each model is normalized by the loss of the PlainVF + Tensor Gauge Field model.  
  Values are shown for several ambient dimensions \(N\).}
  \label{fig:test_loss}
\end{figure}

Figure~\ref{fig:test_loss} shows the corresponding test losses.  
The Tensor Gauge Flow Models again outperform all other model variants across all dimensions \(N\), indicating improved generalization rather than mere overfitting.

\subsubsection{Number of parameters}

\begin{figure}[h!]
  \centering
  \includegraphics[width=0.6\textwidth]{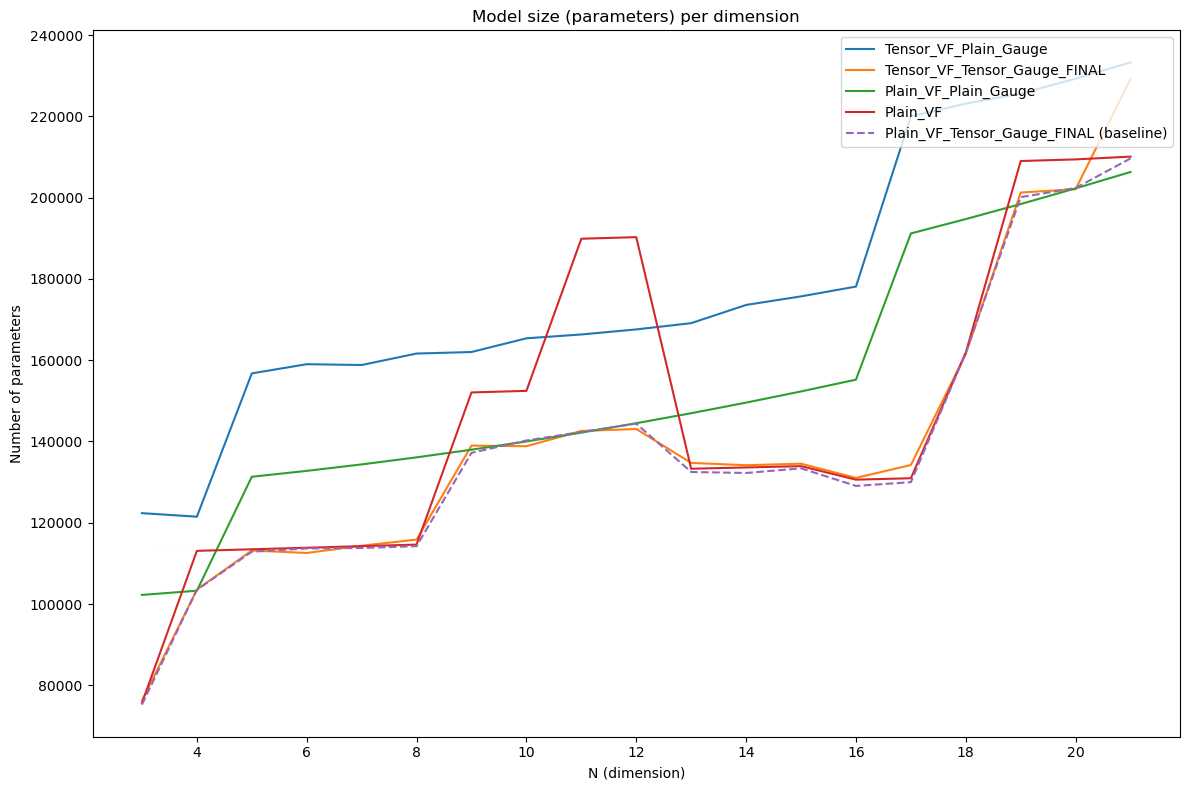}
  \caption{Number of trainable parameters for each model family as a function of ambient dimension \(N\).}
  \label{fig:NumParams}
\end{figure}

As shown in Figure~\ref{fig:NumParams}, the Plain Flow Model uses a slightly larger number of
parameters than the Tensor Gauge Flow Models across all dimensions \(N\).

\newpage

\section{Related Work}
The concept of Tensor Gauge Flow Models simply generalizes both the framework of Gauge Flow Models~\cite{GaugeFlowModels} and Higher Gauge Flow Models~\cite{HigherGaugeFlowModels} from a $1$-form Gauge Field to higher-order Tensor Gauge Fields.  There are several lines of work that are closely related to this construction, spanning gauge-based Flow Models, Riemannian and geometric Flow Matching, and recent high-order Flow Matching formulations.

\paragraph{Gauge Flow Models and Higher Gauge Flow Models.}
Gauge Flow Models (GFM)~\cite{GaugeFlowModels} introduce a learnable Lie-algebra-valued gauge field into the flow ODE, modifying the neural vector field by a gauge correction term that is learned jointly with the base flow.  This yields simulation-free continuous-time generative models trained by Flow Matching and empirically improves performance on Gaussian mixture benchmarks compared to plain Flow Models of similar or larger size~\cite{GaugeFlowModels}.  Higher Gauge Flow Models (HGFM)~\cite{HigherGaugeFlowModels} extend this idea by replacing the underlying Lie algebra with an $L_\infty$-algebra, so that the Higher Gauge Field is an $L_\infty$-algebra-valued differential $1$-form acting on graded vector fields through higher brackets~\cite{HigherGaugeFlowModels}.  This allows one to encode higher group symmetries and higher gauge geometry inside the flow, again with simulation-free Flow Matching training and improved generative performance on synthetic benchmarks.  Tensor Gauge Flow Models can be viewed as a further generalization of GFM and HGFM in which the gauge correction term is constructed from higher-rank Tensor Fields (rather than only $1$-forms) valued in a Lie or $L_\infty$-algebra (or its module), thereby enriching the geometric structure available to the model.

\paragraph{Flow Matching and Riemannian Flow Matching.}
Flow Matching (FM)~\cite{FM} is a simulation-free framework for training continuous-time generative models that directly regresses a velocity field transporting a base distribution to the data distribution along a prescribed family of probability paths~\cite{FM,FlowMatchingGuideCode}.  Instead of optimizing likelihood via continuous normalizing flows \cite{CNFs}, FM constructs a tractable regression objective based on a conditional target vector field and unbiased Monte Carlo estimators, leading to competitive or state-of-the-art generative models across images, audio, and other modalities~\cite{FlowMatchingGuideCode}.  Riemannian Flow Matching (RFM)~\cite{RFM} generalizes FM to flows on Riemannian manifolds by defining suitable premetrics and target vector fields on general geometries; it retains the simulation-free nature of FM while avoiding explicit divergence computations~\cite{RFM}.  RFM and its variants have been used for generative modeling on curved surfaces and meshes~\cite{RFM}, for crystal structure and materials generation (FlowMM)~\cite{FlowMM}, for visuomotor robot policies (Riemannian Flow Matching Policy, RFMP)~\cite{RiemannianFlowMatchingPolicy}, and for structured geometric data such as learned Riemannian representations of shapes and poses~\cite{RFM}.  Tensor Gauge Flow Models adopt the same FM/RFM training paradigm but modify the model vector field by Tensor-valued gauge corrections defined on Fiber Bundles over the data manifold.

\paragraph{Higher-order Flow Matching and High-Order Trajectory Information.}
A growing body of work investigates high-order Flow Matching, where the model is trained not only on first-order velocity fields but also on higher-order derivatives such as accelerations.  Empirically, methods like NRFlow, which jointly learn velocity and acceleration fields via a two-part loss, show improved robustness and smoother transport trajectories in noisy settings~\cite{NRFlow}.  Theoretical work on high-order trajectory refinement proves that higher-order Flow Matching (with detailed analysis for second order) preserves worst-case optimality as a distribution estimator and provides error bounds for refined trajectories~\cite{HighOrderTrajectoryRefinement}.  In parallel, MeanFlow-type approaches have been generalized to \emph{second order}, learning average acceleration fields in addition to average velocities and yielding theoretically grounded high-order flows with efficient one-step sampling~\cite{HighOrderMeanFlow}.  These works demonstrate how FM-style objectives can be lifted from first-order dynamics to higher-order trajectories; Tensor Gauge Flow Models are complementary in that they lift the \emph{state-dependent correction} to higher-order Tensor Gauge Fields on bundles, rather than augmenting the temporal order of the ODE alone.

\paragraph{Flow Matching over Higher-Order Geometric Objects.}
Another direction develops Flow Matching on ``higher-order'' geometric spaces such as spaces of distributions or structured manifolds.  Wasserstein Flow Matching (WFM) lifts FM to the Wasserstein space of probability measures, allowing generative modeling over families of distributions (e.g., shapes or cell populations) by learning flows in distribution space rather than sample space~\cite{WassersteinFlowMatching}.  Metric Flow Matching introduces a metric-aware version of Conditional Flow Matching that interpolates along geodesics in data-dependent metrics, further emphasizing the role of underlying geometry in FM objectives~\cite{MetricFlowMatching}.  These approaches can be interpreted as Flow Matching on higher-level geometric objects (spaces of measures, metric manifolds), while Tensor Gauge Flow Models focus on enriching the \emph{local} geometric structure through Tensor-valued Gauge Fields and higher algebraic data.

\paragraph{Gauge- and Geometry-aware Neural Architectures.}
Finally, Tensor Gauge Flow Models are related in spirit to gauge-equivariant and geometry-aware neural architectures.  Gauge-equivariant convolutional networks generalize group-equivariant CNNs to local gauge transformations on manifolds, using gauge connections and differential forms to define consistent convolutions on curved domains and spherical meshes~\cite{CohenGaugeCNN,HussainGaugeCNNdMRI}.  Riemannian and geometric deep learning methods more broadly exploit manifold structure, vector fields, and differential forms to define intrinsic operations on curved spaces~\cite{MontiMoNet,BronsteinGeometricDL}.  Unlike these architectures, which encode gauge or geometric structure at the level of layer design, Gauge Flow Models, Higher Gauge Flow Models, and Tensor Gauge Flow Models integrate gauge-theoretic structure directly into the continuous-time generative dynamics via gauge-corrected vector fields trained with Flow Matching or Riemannian Flow Matching.

\noindent In summary, Tensor Gauge Flow Models extend the Gauge Flow and Higher Gauge Flow frameworks to higher-order Tensor Gauge Fields, while remaining compatible with the Flow Matching and Riemannian Flow Matching objectives that underpin many recent geometric and high-order generative models.

\newpage
\printbibliography

\end{document}